\documentclass[10pt,twocolumn,letterpaper]{article}

\usepackage{cvpr}              

\usepackage{graphicx}
\usepackage{amsmath}
\usepackage{amssymb}
\usepackage{booktabs}

\usepackage{multirow}
\usepackage{adjustbox}
\usepackage{multirow}
\usepackage{adjustbox}
\usepackage[ruled,vlined,linesnumbered]{algorithm2e}
\usepackage{algorithmic}
\usepackage{xcolor, colortbl}
\definecolor{grey}{rgb}{0.4,0.4,0.4}

\usepackage[pagebackref,breaklinks,colorlinks]{hyperref}

\usepackage[capitalize]{cleveref}
\crefname{section}{Sec.}{Secs.}
\Crefname{section}{Section}{Sections}
\Crefname{table}{Table}{Tables}
\crefname{table}{Tab.}{Tabs.}

\def\cvprPaperID{*****} 
\def\confName{CVPR}
\def\confYear{2022}

\begin{document}

\title{Empirical Evaluation of Data Augmentations for Biobehavioral Time Series Data with Deep Learning}

\author{Huiyuan Yang, Han Yu and Akane Sano\\
Department of Electrical Computer Engineering\\
Rice University, Houston TX 77005, USA\\
{\tt\small  \{hy48, hy29, Akane.Sano\}@rice.edu}
}
\maketitle

\begin{abstract}
Deep learning has performed remarkably well on many tasks recently. 
However, the superior performance of deep models relies heavily on the availability of a large number of training data, which limits the wide adaptation of deep models on various clinical and affective computing tasks, as the labeled data are usually very limited.
As an effective technique to increase the data variability and thus train deep models with better generalization, data augmentation (DA) is a critical step for the success of deep learning models on biobehavioral time series data. However, the effectiveness of various DAs for different datasets with different tasks and deep models is understudied for biobehavioral time series data.
In this paper, we first systematically review eight basic DA methods for biobehavioral time series data, and  evaluate the effects on seven datasets with three backbones. 
Next, we explore adapting more recent DA techniques (\textit{i.e., automatic augmentation, random augmentation}) to biobehavioral time series data by designing a new policy architecture applicable to time series data.
Last, we try to answer the question of why a DA is effective (\textit{or not}) by first summarizing two desired attributes for augmentations (\textit{challenging} and \textit{faithful}), and then utilizing two metrics to quantitatively measure the corresponding attributes, which can guide us in the search for more effective DA for biobehavioral time series data
 by designing more challenging but still faithful transformations. 
 Our code and results are available at \href{https://github.com/comp-well-org/Data_Augmentation_for_Biobehavioral_Time_Series_Data.git}{Link}.
\end{abstract}

\section{Introduction}
\label{sec:intro}

Deep learning performs remarkably well in many fields, including computer vision (CV), natural language processing (NLP), and recently time series-related tasks \cite{ismail2019deep, wen2020time, gamboa2017deep}. 
Those successful applications increasingly inspire researchers to embrace deep learning for solving issues in human-centered applications that use physiological and behavioral time series data.
However, the superior performance of deep models relies heavily on the availability of a large number of training data, but
unfortunately, many human centered applications (\textit{i.e., healthcare tasks}) usually do not have enough labeled samples, which may limit the wide adaptation of deep models to various computing tasks.

As an effective technique to increase the data variability and thus train deep models with better generalization, data augmentation (DA) is a critical step for the successful applications of deep learning models.
While DA can yield considerable performance improvements, they do require domain knowledge and are task- and domain-dependent. 
For example, image rotation, a likely class-preserving behavior, is designed to rotate the input by some number of degrees. 
The image's class can still be recognized by humans, thus allowing the model to generalize in a way humans expect it to generalize. 
However, such an effective random angle-based rotation operation may not be applicable to other domains, \textit{i.e., wearable data}.
In addition, searching for the most effective DA methods for a new dataset is very time-consuming, and this motivated the proposal of several automatic DA search algorithms \cite{cubuk2019autoaugment,  lim2019fast, cubuk2020randaugment,  li2020dada, liu2021direct}.
 \begin{figure*}[!htp]
  \centering
  \includegraphics[width=1.0\linewidth]{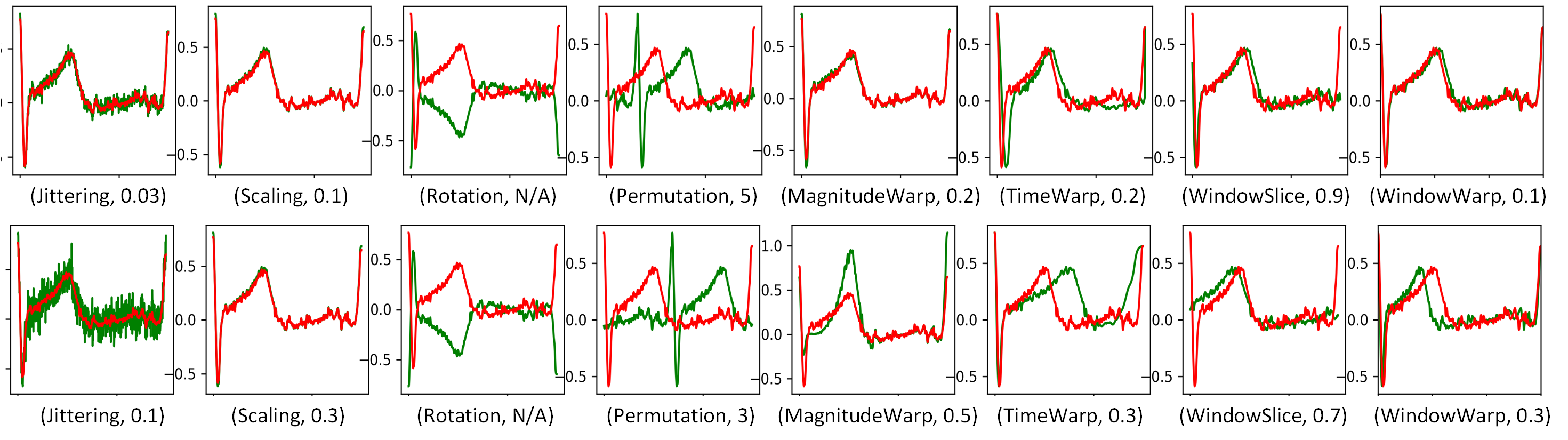}
  \caption{Examples of different data augmentation methods used in the experiments. The red lines indicate the input data and the green lines are the augmented data based on the eight data augmentation operations including \textit{Jittering, Scaling, Rotation, Permutation, Magnitude Warping, Time Warping, and Window Warping} with two different magnitudes.}
  \label{fig:DA_examples}
\end{figure*}

The existing DA literature mainly focuses on computer vision, but its application to other domains, \textit{i.e, biobehavioral time series data}, is understudied.
A few works investigated the effectiveness of basic DA methods for time series and wearable data \cite{um2017data, iwana2021empirical, wen2020time, alawneh2021enhancing}.
However, those works only investigated the very basic DAs, leaving the more recent DA techniques (\textit{i.e., automatic DA}) unexplored. 
More importantly, it is still an open question of why a DA method works, and how to quantify its effectiveness.
Therefore, in this paper, we first systematically review various basic DA methods for biobehavioral time series data,  evaluating the effects on different datasets with varied backbones and tasks. 
Next, we validate the effectiveness of adapting  more recent DA techniques (\textit{i.e., automatic DA}) to biobehavioral time series data.
Following the DADA\cite{li2020dada}, we designed a different policy architecture where the operations are differentiable with respect to different time series DA methods. Therefore, the model can be applied to biobehavioral time series data, and the DA parameters and deep model weights can be jointly optimized.
Lastly, we try to answer the open question of why a DA works(\textit{or not}), by first summarizing two desired attributes (\textit{challenging} and \textit{faithful}) for an effective DA, and then utilizing two metrics to quantitatively measure the two attributes.  We find that an effective DA needs to generate challenging but still faithful transformations, which can guide us for the search of more effective DA for biobehavioral time series data.
The contributions of this work are summarized as follows:
\begin{itemize}
     \item  A comprehensive and systematic evaluation of \textbf{eight} data augmentation methods on \textbf{seven} biomedical time series datasets with \textbf{three} backbones for different tasks.
     \item We revisit the automatic DA methods to make the operations are differentiable with respect to different time series DA methods, therefore can be applied to biobehavioral time series data. Besides, random DA is also investigated to boost efficiency. 
     \item  We summarize two desired attributes for an effective DA, and adopt two metrics to quantitatively measure the two attributes respectively. Recommendations are summarized for the search of more effective data augmentation methods.
\end{itemize}

 \section{Related Works}

\subsection{Augmentations for Biobehavioral Time Series Data}
Most of the basic DA methods are borrowed or inspired from image or time series data augmentation, such as flipping, cropping and noise addition. These augmentation methods rely on adding random transformations to the training data.
Um et al. \cite{um2017data} systematically evaluated six DA methods for wearable sensor data based Parkinson's disease monitoring, and found that the combination of rotational and permutational data augmentation methods improve the baseline performance the most.
Ohashi et al. \cite{ohashi2017augmenting} proposed a rotation based data augmentation method for wearable data, which can take the physical constraint into account.
Alawneh et al. \cite{alawneh2021enhancing} investigated the benefits of adopting time series data augmentation methods to biomedical time series data, and demonstrated the improved accuracy of several deep learning models for human activity recognition.
Eyobu and Han \cite{steven2018feature} proposed an ensemble of feature space augmentation methods, which was used for human activity classification based on wearable  inertial measurement unit (IMU) sensors.
Besides, DA methods have been also used to balance the dataset. For example, Cao et al. \cite{cao2020novel} used DA methods to balance the number of samples among different categories for automated heart disease detection.
However, those related works only investigated the very basic DAs, and the effectiveness of adapting  more advanced DAs is not explored yet for biobehavioral time series data.
More importantly,  the previous works did not explore the question why a DA is effective, and vice versa.


\subsection{Automatic Data Augmentation}
DA can be very useful for the training of deep models, but the success relies heavily on domain knowledge and also the extensive experiments to select the effective DA policies for a target dataset. Otherwise, a model may be negatively impacted by some DA policies \cite{iwana2021empirical, wen2020time}. Therefore, it is nontrivial and desired to select the effective DA policy for a new dataset automatically. 
The goal of automatic data augmentation is to search for effective data augmentation policies that, when applied during the model training, will minimize its validation loss, therefore better generalization ability. 
The pioneering work, AutoAugment \cite{cubuk2019autoaugment}, formats the process of searching DA as an optimization problem to search for the parameters of augmentation,
and follwing work \cite{ho2019population, lim2019fast, li2020dada, liu2021direct, cubuk2020randaugment} were later proposed to improve the efficiency. 
Despite the success of automatic DA for computer vision tasks, the adaption to biobehavioral time series data is understudied.
The only work we know is \cite{rommel2021cadda}, which investigated the automatic differentiable data augmentation for EEG signals. However, our work is different with \cite{rommel2021cadda}, as we target more diverse types of data and  DA methods, and more importantly, we explore to explain why a DA works or not.

\begin{figure}[!ht]
  \centering
  \includegraphics[width=1.0\linewidth]{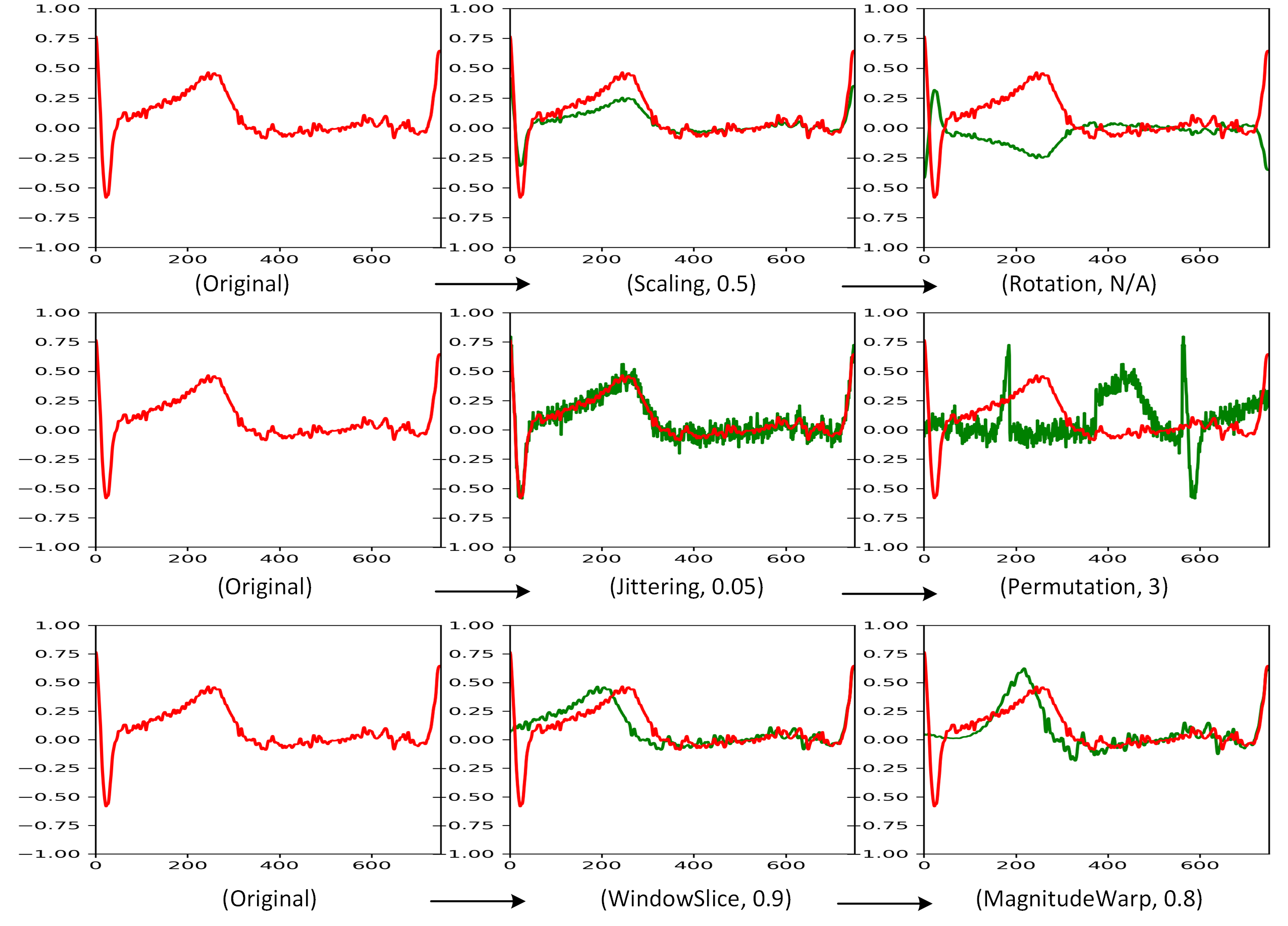}
  \caption{Examples of time series data augmented by two consecutive operations. The first column is the original input signal, with two random operations sequentially applied to the input signal, the final augmented signal is shown in the third column (as green line). Note that, not only the name of operations and magnitude matter, but also the order of operation.  }
  \label{fig:DA_randAugment}
\end{figure}

\subsection{Quantitative Measurement of Effectiveness for Data Augmentation}
Although the effectiveness of DA is well acknowledged, a quantitative evaluation of the effectiveness of DA is still an open question. Currently, the most well-known hypothesis is that effective DA can produce samples from an "overlapping but different" distribution \cite{bengio2011deep, liu2021divaug}, therefore improving generalization by training with the diverse samples. However, the role of distribution shift in training remains unclear.
A more recent work \cite{gontijo2020tradeoffs} studied to quantify how DA improves model generalization, and introduced two measures: \textit{Affinity} and \textit{Diversity}, to predict the performance of an augmentation method. During our experiments, we  adopt those two metrics to jointly evaluate the two attributes of different augmentations.

\section{Methods}
We performed extensive experiments with various DA methods on different datasets with different tasks and backbones. After that, an automatic DA search strategy is adapted to jointly optimize the deep models and also the best DA policies for wearable data.
To avoid the complicated searching procedure of automatic DA while keeping the effectiveness, a random augmentation procedure was then adapted in our experiments.
Lastly, We also explored to answer the open questions of why  some DA methods are more effective than others by quantitatively measuring the effectiveness.

\subsection{Basic Data Augmentation Methods}
DA methods rely on adding random transformations to the training data, and the transformations can be generally classified into three categories:  magnitude-based, time-based, and frequency-based transformations.  Magnitude-based transformations are applied to the wearable data along the variate or value axes. Time-based transformations change the time steps, and frequency-based transformations warp the frequencies, respectively. During our experiments, we will mainly focus on the magnitude and time-based transformations.

Let $\mathcal{T}_{\alpha}$ denote an augmentation operations parameterized by $\alpha$, given input data $x$, this procedure outputs augmented data $\hat{x} = \mathcal{T}_{\alpha}(x)$, where $\alpha$ controls the magnitude of the operation $\mathcal{T}$.
We consider a set of  DA methods drawn from the traditional time series processing literature. Specifically, our augmentation set consists of \textbf{eight}  operations: \textit{Jittering, Scaling, Rotation, Permutation, Magnitude Warping, Time Warping, Window Slicing, and Window Warping}. 

Given an input data $\mathbf{x} = [x_{1}, x_{2}, \dots, x_{T}]$ with $T$, the number of time steps, and each element $x_{t}$ can be univariate or multivariate. 

\noindent
\textbf{Jittering.} A random noise is added to the input data:
\begin{equation}
    \mathbf{\hat{x}} = \mathcal{T}_{\alpha}(\mathbf{x}) = [x_{1} + \epsilon_{1}, x_{2} + \epsilon_{2}, \dots, x_{T} +\epsilon_{T} ]
\end{equation}
where $\mathbf{\hat{x}}$ is the augmented data, $\epsilon$ is a random noise (\textit{i.e.,} Gaussian noise) added to each time step $t$. Assume $\epsilon \sim \mathcal{N}(0, \sigma^2)$, and the standard deviation $\alpha = \{\sigma \}$ of the added noise is a hyper-parameter that needs to be pre-determined. 

\noindent
\textbf{Scaling.} The magnitude of the data in a window is changed by multiplying a random scalar. 
\begin{equation}
    \mathbf{\hat{x}} = \mathcal{T}_{\alpha}(\mathbf{x}) = [\epsilon x_{1}, \epsilon x_{2}, \dots, \epsilon x_{T} ]
\end{equation}
where $\epsilon$ can be determined by a Gaussian distribution $\epsilon \sim \mathcal{N}(1, \sigma^2)$ with $\alpha = \{\sigma \}$ as a hyperparameter.


\noindent
\textbf{Rotation.} Rotate each element by a random rotation matrix. Although rotating data by a random angle can create plausible patterns for images, it might not be suitable for time series data. A widely used alternative is flipping, which is defined as:
\begin{equation}
    \mathbf{\hat{x}} = \mathcal{T}_{\alpha}(\mathbf{x}) = [-x_{1}, -x_{2}, \dots, -x_{T}]
\end{equation}

\noindent
\textbf{Permutation.} Perturb the location of the data in a single window.
It should be noted that permutation operation does not preserve time dependencies. Permutation can be performed in two ways: equal sized segments and variable sized segments.  First, the data $\mathbf{x}$ is split into $N$ segments, each of the segment has a length of $\frac{T}{N}$ (\textit{assuming equal sized segments});  then the location of those segments are randomly permuted.
\begin{equation}
    \mathbf{\hat{x}} = \mathcal{T}_{\alpha}(\mathbf{x}) = [segment_{1^{*}}, \dots ,segment_{N^{*}} ]
\end{equation}
where $segment_{i^{*}}$ is the $\operatorname{i^{*}-th}$ segment of the input data $\mathbf{x}$, and $i^{*} \in [1, N]$. The number of segments $\alpha = \{ N\}$ is a hyperparameter to be pre-determined.

\noindent
\textbf{Magnitude Warping.} Change the magnitude of each sample by convolving the data window with a smooth curve. 
\begin{equation}
    \mathbf{\hat{x}} = \mathcal{T}_{\alpha}(\mathbf{x}) = [\gamma_{1} x_{1}, \gamma_{2} x_{2}, \dots, \gamma_{T} x_{T} ]
\end{equation}
where $\gamma_{1}, \gamma_{2}, \dots \gamma_{T}$ is a sequence created by interpolating a cubic spline function $\mathcal{S}(\mathbf{u})$ with knots $\mathbf{u} = u_{1}, u_{2}, \dots u_{I}$. $I$ is the number of knots and each knot $u_{i}$ is sampled from a Gaussian distribution $\mathcal{N}(1, \sigma^2)$, therefore, the operation parameters $\alpha = \{\sigma, I\}$.

\noindent
\textbf{Time Warping.} Perturb the temporal location by smoothly distorting the time intervals between samples, which is similar to the magnitude warping operation and defined as:
\begin{equation}
    \mathbf{\hat{x}} = \mathcal{T}_{\alpha}(\mathbf{x}) = [ x_{\gamma_{(1)}},  x_{\gamma_{(2)}}, \dots,  x_{\gamma_{(T)}} ]
\end{equation}
where $\gamma(\cdot ): i \rightarrow j, and \ i, j \in [1, T]$ is a warping function that warps the time steps based on a smooth curve. The smooth curve is defined by a cubic spline $\mathcal{S}(\mathbf{u})$, which is exactly the same as used in the magnitude warping operation, and the operation parameters $\alpha = \{\sigma, I\}$.

\noindent
\textbf{Window Slicing.} Slice time steps off the ends of the pattern, which is equivalent to cropping for image data augmentation. The operation is defined as:
\begin{equation}
    \mathbf{\hat{x}} = \mathcal{T}_{\alpha}(\mathbf{x}) = [ x_{\delta}, \dots,  x_{t}, \dots, x_{W + \delta} ]
\end{equation}
where $0 \leq W \leq T$ is the size of a window that needs to be pre-determined, and  $\delta$ is a random integer such that $ 0 \leq \delta \leq T- W $, and the operation parameters $\alpha = \{W \}$.

\noindent
\textbf{Window Warping.} Randomly select a window with the length of $W$ from the time series and stretch it by $K$ or contract it by $\frac{1}{K}$. Linear interpolation is used for other part of the time series, so that the output will have equal length to the input. Note that, in this paper we only consider $K=2$, but  other ratios could be used as well. The length of window is a hyper-parameter to be pre-determined and $\alpha = \{W \}$

\vspace{1mm}
With different operations and their corresponding magnitude parameters $\alpha$, the generated augmented data are also different. As shown in Fig.\ref{fig:DA_examples}, the two rows represent the examples of the augmented data where the eight basic DA methods are applied to the same input with two different magnitude parameters $\alpha$ for each operation.
Besides, different DAs can be combined together to generate more diverse augmented data, as shown in Fig.\ref{fig:DA_randAugment}. As a result, manually searching for the optimal DA(s) can be very challenging and time consuming.

\subsection{Advanced Data Augmentation Methods}

\subsubsection{Automatic Data Augmentation}
To alleviate the computation burden, differentiable automatic DA \cite{li2020dada} was proposed to greatly improve the optimization efficiency through relaxing the optimization process as differentiable and jointly optimized DA parameters with deep model weights.
In this paper, we revisit the idea of automatic DA \cite{li2020dada, rommel2021cadda} to make the operations differentiable with respect to different time series DA methods. Therefore, the model can be applied to biobehavioral time series data, and the weights of deep models and  augmentation parameters can be jointly optimized.

A collection of DA policies contains $K$ sub-policies $\mathbf{P} = [p_1, p_2, \dots p_{K}]$, and each sub-policy $p_{k}$ includes $J$ basic DA operations (\textit{i.e., jittering, scaling, rotation}) that are applied to input signal sequentially. Each operation can be represented as $\mathbf{O}_{k}^{i}\big(\mathbf{x}; p_{k}^{i}, m_{k}^{i} \big)$, $k \in [1, K] $, $i \in [0, J] $, where $K$ is the total number of sub-policies, $J$ is the operations included in each sub-policy. $ p_{k}^{i}$ is the probability of applying the operation and $m_{k}^{i}$ represents the magnitude of the corresponding operation.
The objective is to jointly optimize a model's parameter $\theta$ and also the augmentation parameters $\alpha$, where $\alpha$ represents both the probability $p_{k}^{i}$ and magnitude $m_{k}^{i}$. Therefore, we can not only train a deep model that generalizes well on the testing dataset, but also learn the optimal augmentation policies.

\begin{figure*}[!htp]
\vspace{-5mm}
  \centering
  \subfloat[PTB-XL]{\includegraphics[width=0.33\linewidth]{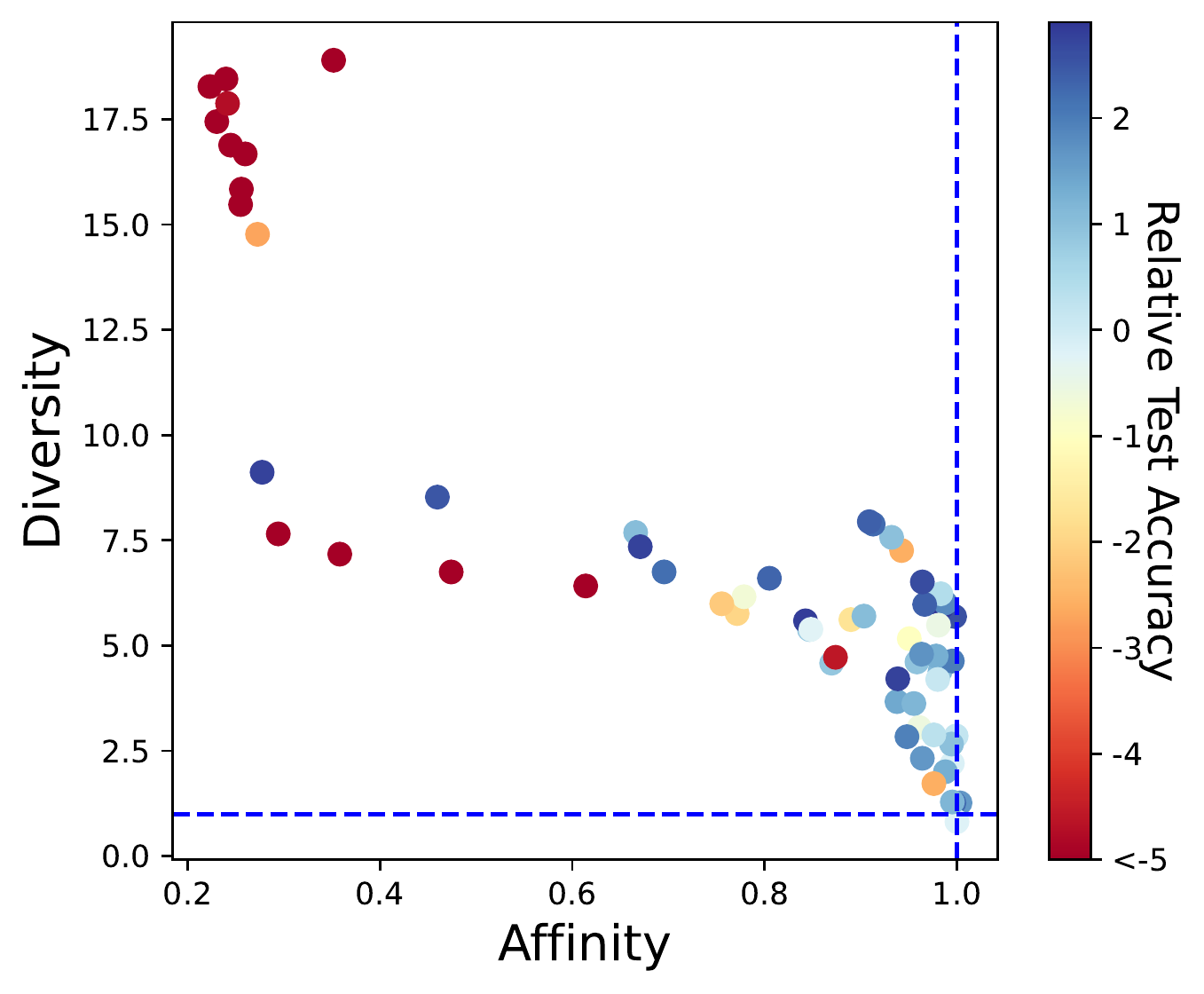}}
  \subfloat[PAMAP2]{\includegraphics[width=0.33\linewidth]{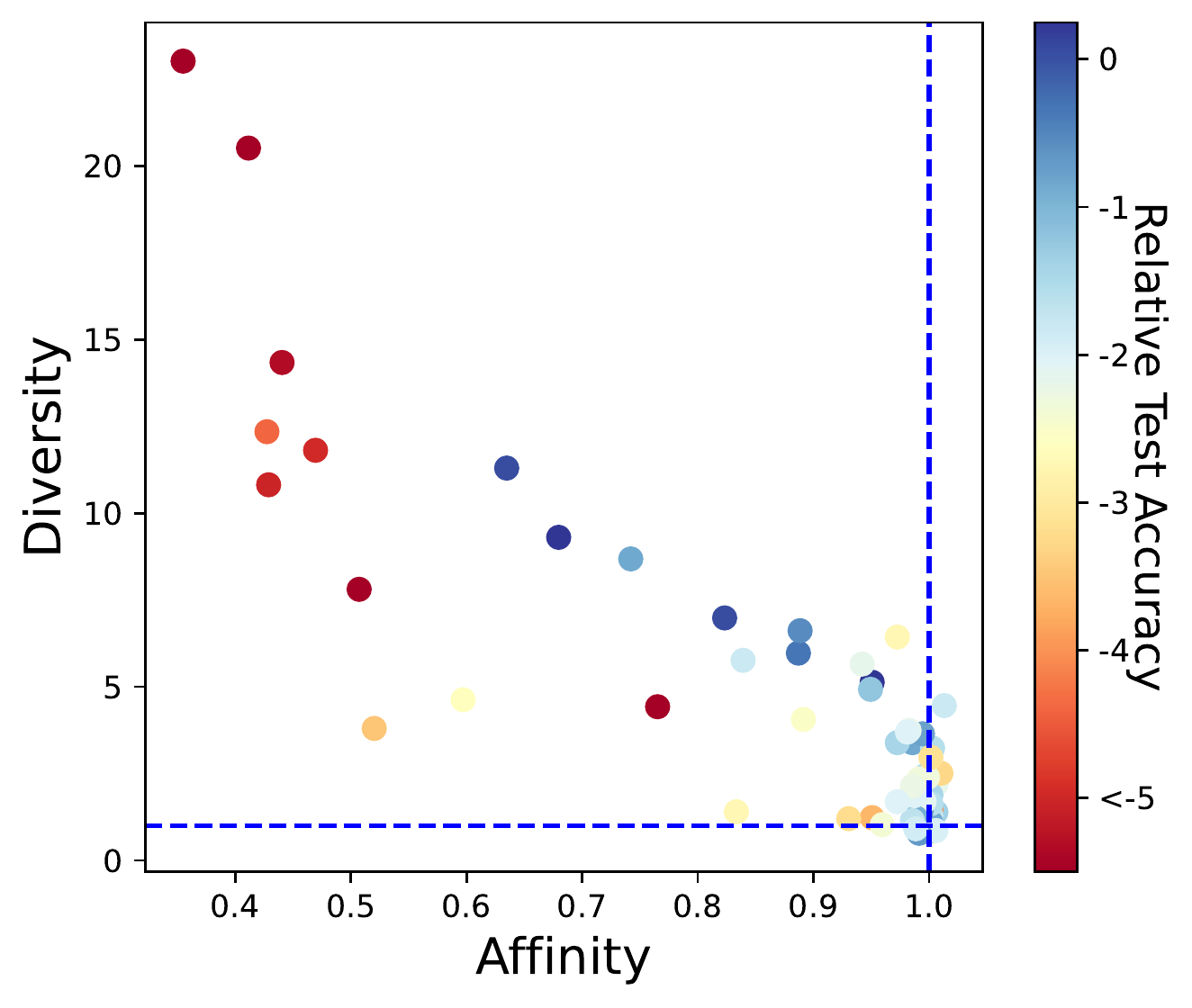}}
  \subfloat[Illustration]{\includegraphics[width=0.28\linewidth]{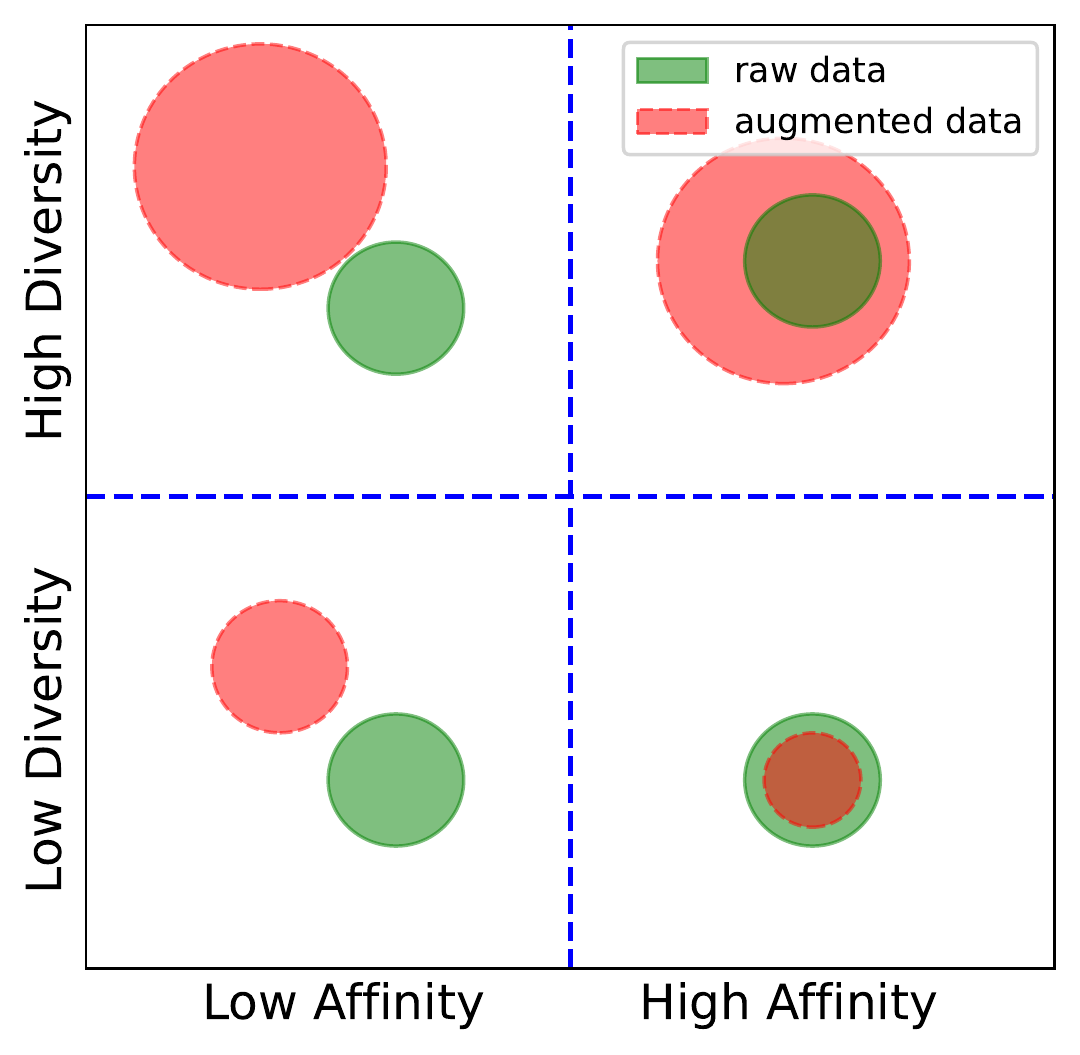}}
  \caption{Augmentation performance is determined by both affinity and diversity. Test accuracy plotted against each of Affinity and Diversity in the PTB-XL dataset (a) and PAMAP2 dataset (b), where each point represents  a different DA (65 augmentations in total).  Color shows the final test accuracy relative to the baseline model trained without augmentation. (c) Illustration of how raw data and augmented data are associated in terms of the two metrics.  A larger circle represents higher diversity while the distributional similarity is depicted trough the overlap of circles.  Test accuracy generally improves with both high affinity and high diversity (\textit{upper right space}). }
  \label{fig:quantitative_evaluation}
\end{figure*}

\subsubsection{Practical Data Augmentation}


In spite of the benefits of learned DA policies from the target dataset, the computational requirements as well as the added complexity of the jointed optimization procedures can be prohibitive.
Therefore, we also deployed random augmentation, which eliminate the searching phase but still keep the benefits of different augmentations. More specifically, 
we follow RandAugment \cite{cubuk2020randaugment} to replace the differentiable automatic DA module with a parameter-free procedure for biomedical time series data, which always selects an operation with uniform probability $\frac{1}{K}$, where $K$ is the total number of operations. 
The algorithm can be defined as: $randaugment(J, M)$, where $J$ is the number of augmentation transformations to be selected from the $K$ operation pool, $M$ is the magnitude for all the transformations, then the $J$ operations with magnitude $M$ will be sequentially applied to the input data. 
With this method, the DA procedure can be easily plugged in the training of a deep model without requiring special attention or designation.

\begin{table*}[ht!]
\caption{Performance in terms of 8 different DA methods and three backbones (\textit{MLP, Conv-1D and ResNet-1D}) are reported on night datasets. Bold numbers indicate the best performance.
The top three most effective DA methods (\textit{if exists}) that improve the baseline for different datasets and backbones are colored in gray, and the grayscale represents the corresponding improvements.
}
\label{table_different_DA}
\begin{center}
\begin{adjustbox}{max width=\textwidth}
\small
\begin{tabular}{lllllllllll}
\toprule
Dataset  & Backbone  & \multicolumn{1}{l}{No Aug} & \multicolumn{1}{l}{Jittering} & \multicolumn{1}{l}{Scaling} & \multicolumn{1}{l}{Rotation} & \multicolumn{1}{l}{Permutation} & \multicolumn{1}{l}{\begin{tabular}[c]{@{}l@{}}Magnitude\\ Warp\end{tabular}} & \multicolumn{1}{l}{\begin{tabular}[c]{@{}l@{}}Time\\ Warp\end{tabular}} & \multicolumn{1}{l}{\begin{tabular}[c]{@{}l@{}}Window\\ Slice\end{tabular}} & \begin{tabular}[c]{@{}l@{}}Window\\ Warp\end{tabular}  \\ \midrule

                                                                                                                            
                                                                                                                             
\multirow{3}{*}{PTB-XL}       & MLP       & 60.29&	58.66&	61.14&	55.45&	\cellcolor{grey!35}64.04&	60.23&	60.41&\cellcolor{grey!35}\textbf{66.34}&	\cellcolor{grey!15}63.80        \\ \cline{2-2} 
                              & Conv-1D   & 75.00&  74.88&\cellcolor{grey!35}\textbf{77.91}&	71.55&	\cellcolor{grey!15}76.63&	76.57&	64.41&	75.91&	\cellcolor{grey!25}77.72              \\ \cline{2-2} 
                              & ResNet-1D & 77.24&	\cellcolor{grey!25}77.60&\cellcolor{grey!35}\textbf{78.03}&	63.98&	76.57&	76.51&	67.07&	75.48&	75.67       \\ \midrule
                                                                                                                    
\multirow{3}{*}{Apnea-ECG} & MLP       & 53.57&53.46&54.76&54.63&55.26&54.42&\cellcolor{grey!35}\textbf{57.90}&\cellcolor{grey!15}56.19&\cellcolor{grey!25}56.98                  \\ \cline{2-2} 
                              & Conv-1D   & 80.14&80.00&73.19&\cellcolor{grey!35}\textbf{82.67}&\cellcolor{grey!25}81.27&71.76&78.54&75.94&79.20               \\ \cline{2-2} 
                              & ResNet-1D & 76.64&76.44&70.64&\cellcolor{grey!35}\textbf{77.74}&\cellcolor{grey!25}77.26&70.19&76.28&74.32&76.34               \\ \midrule
                                                                                                                   
 \multirow{3}{*}{Sleep-EDFE} & MLP       & 49.85&50.59&50.69&52.36&\cellcolor{grey!35}\textbf{56.27}&50.66&47.76&\cellcolor{grey!25}53.78&\cellcolor{grey!15}53.01                \\ \cline{2-2} 
                              & Conv-1D   & 83.62&83.59&84.31&\cellcolor{grey!15}84.56&\cellcolor{grey!35}\textbf{85.31}&\cellcolor{grey!25}85.00&82.80&81.12&84.30               \\ \cline{2-2} 
                              & ResNet-1D &83.61&81.55&83.40&\cellcolor{grey!25}84.10&\cellcolor{grey!35}\textbf{85.01}&83.09&82.09&82.80&\cellcolor{grey!15}83.98                 \\ \midrule
                                                                                                                    
\multirow{3}{*}{MMIDB-EEG}    & MLP       &77.24&	\cellcolor{grey!15}78.10&	74.76&	50.92&	68.39&	73.14&	\cellcolor{grey!25}78.21&	74.43&	\cellcolor{grey!35}\textbf{78.68}  \\ \cline{2-2} 
                              & Conv-1D   &79.29&	\cellcolor{grey!35}\textbf{79.72}&	77.13&	48.98&	75.62&	77.67&	77.67&	74.97&	77.99       \\ \cline{2-2} 
                              & ResNet-1D &76.16&	\cellcolor{grey!15}79.07&	\cellcolor{grey!35}\textbf{80.15}&	49.30&	69.58&	74.54&	\cellcolor{grey!25}79.61&	70.98&	77.99       \\ \midrule                            
                                                                                                                    
\multirow{3}{*}{CLAS}         & MLP   & 76.71&75.37&74.51&77.02&\cellcolor{grey!35}\textbf{79.06}&76.16&\cellcolor{grey!25}78.59&\cellcolor{grey!15}77.88&77.49                     \\ \cline{2-2} 
                              & Conv-1D   & 62.83&66.12&70.59&\cellcolor{grey!25}76.54&73.80&71.37&\cellcolor{grey!35}\textbf{77.96}&\cellcolor{grey!15}75.37&73.64                         \\ \cline{2-2} 
                              & ResNet-1D & \textbf{76.54}&71.29&66.82&70.04&71.84&69.10&74.47&66.59&70.18                         \\ \midrule
                                                                                                                             
\multirow{3}{*}{PAMAP2}       & MLP     & 61.15&	61.45&	60.86&	13.88&	\cellcolor{grey!35}\textbf{66.17}&	61.74&	59.82&	\cellcolor{grey!15}62.92&	\cellcolor{grey!25}65.14            \\ \cline{2-2} 
                              & Conv-1D   & 89.22&	\cellcolor{grey!25}91.58&	91.14&	67.06&	\cellcolor{grey!25}91.58&	\cellcolor{grey!35}\textbf{92.02}&	88.33&	\cellcolor{grey!15}91.29&	89.07         \\ \cline{2-2} 
                              & ResNet-1D & 89.81&	86.85&	88.18&	65.14&	85.82&	\cellcolor{grey!35}\textbf{92.02}&	88.63&	88.33&	87.59          \\ \midrule
                                                                                                                   
\multirow{3}{*}{UCI-HAR}      & MLP      &87.89&	87.82&	86.33&	19.41&	\cellcolor{grey!35}\textbf{90.80}&  85.85&	82.86&  \cellcolor{grey!25}89.96&  \cellcolor{grey!15}89.68         \\ \cline{2-2} 
                              & Conv-1D  &91.92&	89.11&	91.79&	73.91&	\cellcolor{grey!35}\textbf{93.55}&	90.19&	89.41&	91.75&	\cellcolor{grey!25}92.20            \\ \cline{2-2} 
                              & ResNet-1D&89.79&	87.95&	88.63&	65.12&	\cellcolor{grey!25}91.89&	89.62&	88.39&	\cellcolor{grey!15}90.91&	\cellcolor{grey!35}\textbf{92.94}          \\ 
                              \bottomrule                            
\end{tabular}
\end{adjustbox}
\end{center}
\vspace{-5mm}
\end{table*}

\subsection{Quantitative Measurement for Data Augmentation}
We explore to answer the open questions of why  some DA methods are more effective, and vice versa. Following previous work\cite{tamkin2020viewmaker, gontijo2020tradeoffs}, we first summarize the desired attributes (\textit{challenging} and \textit{faithful})for effective DA, and then we adopt the metrics to quantitatively measure the two attributes respectively. 
Specifically, \textit{challenging} refers to the attribute that the augmented data should be challenging for a deep model to over-fit, which
is defined as the ratio of final training loss of a model trained with a given augmentation, and the loss of the model trained on original data.
\textit{Faithful} attribute requires the augmented data should not be so strong that make the learning task impossible, which
is defined as the ratio between the validation accuracy of a model trained on clean data and tested on  an augmented validation set, and the accuracy of the same model tested on clean data.

\section{Experiments}
Through the experiments, we aim to answer the following research questions:

\textbf{R1}: \textit{What is the most effective DA method for a given backbone and dataset?}

\textbf{R2}: \textit{what are the factors that impact the selection of DA?}

\textbf{R3}: \textit{What are the general conclusion we could make for DA methods in biobehavioral time series data?}

\textbf{R4}: \textit{Why are some DA methods more effective than the others?}

\subsection{Datasets and Implementation Details}
\textbf{Datasets.}
We conduct extensive experiments on \textbf{seven} biomedical time series datasets, including
electrocardiogram (ECG) data: PTB-XL \cite{wagner2020ptb} and Apnea-ECG \cite{penzel2000apnea};
electroencephalogram (EEG) data: Sleep-EDF (expanded) \cite{kemp2018sleep} and MMIDB-EEG \cite{Schalk2004};
electrodermal activity (EDA) data: CLAS \cite{markova2019clas};
inertial measurement unit (IMU) data: PAMAP2 \cite{reiss2012introducing} and UCI-HAR \cite{reyes2016transition}.
A detailed description of those datasets can be find in Appendix \ref{appendix:dataset}.


\textbf{Implementation Details.} 
We use three deep learning architectures, including MLP, Conv-1d, and ResNet-1d. These networks were chosen due to being effective in time series data and also being used in a wide range of biomedical applications.

We use an Adam optimizer with initial learning rate of $1\times10^{-3}$, and the learning rate is decayed by $0.9$ after every $5$ epoches. The batch size is 100, and we train the model for 50 epochs.
Our model is implemented in the PyTorch deep learning framework, and is trained and tested on the NVIDIA GeForce 3090Ti GPU.  To evaluate the performance, average accuracy of three runs is reported.  
For fair comparison, we use the default value as magnitude for different basic augmentations on different datasets, therefore we can guarantee that the varied performance is caused by augmentation rather than fine-tuning. 
For random DA, we set $J=2$, $K=8$ and magnitude $M=12$ during our experiments.
For automatic DA, $K=14$ sub-policies are randomly generated from the 8 basic DA methods, and each sub-policy contains $J=2$ operations.


\subsection{Experimental Results}

\subsubsection{Performance of eight basic DA methods.}
Experiments were first conducted on seven datasets with three different (\textit{MLP, Conv-1D and ResNet18-1D}) backbones and eight basic DA methods,
(\textit{Jittering, Scaling, Rotation, Permutation, Magnitude Warp, Time Warp, Window Slice, Window Warp}). From the results in Table.\ref{table_different_DA}, We may find that:

I).\textit{ All the Tasks can benefit from DAs.} (\textbf{R1,R3}) Comparing with the models trained \textit{w/o}  augmentation, training with proper DA usually can achieve higher performance, and the improvement ranges from 0.5\% to 15\% over different datasets,
demonstrating the effectiveness of DA methods for biobehavioral signals. It is worth to note that although augmentations are generally beneficial, the effectiveness varies over different datasets and backbones.

II). \textit{The effectiveness of DA depends on many factors.} (\textbf{R2})
From the grey colored areas, our first impression is that there is no such a single augmentation method works equally well for all the different datasets.
For example, the \textit{permutation}  shows improved or comparable performance on most of the datasets, but decreased performance is also observed for the MMIDB-EEG dataset. 
The dataset varies in terms of data type (\textit{e.g., ECG, EEG, EDA and IMU}) and tasks (\textit{sleep quality, heart disease, human activities, etc}).
We can find  that the effectiveness of an augmentation not only depends on the dataset itself (\textit{data type, task}), but also the selection of backbones.
For example, with the same backbone (\textit{i.e., ResNet-1D}) and data type (\textit{ECG}), the top three augmentations for PTB-XL and Apnea-ECG are totally different. \textit{scaling} and \textit{jittering} work the best for PTB-XL dataset, while \textit{rotation} and \textit{permutation} show the highest improvement for the Apnea-ECG dataset. 
In addition to how different tasks impact the effectiveness of different augmentations, the backbone also effect the choice of DA methods.
For example, in the MMIDB-EEG dataset,  \textit{Window warp} reports the highest performance for the \textit{MLP} backbone, while \textit{Jittering} and \textit{Scaling} achieve the best performance for \textit{Conv-1D} and \textit{ResNet-1D} backbone respectively. 

Besides, we also observe that \textit{permutation} appears 13 times in the top three augmentations, which counts for around half of the rows. Therefore, it is suggested to try \textit{permutation} first for the related applications.

\begin{table}[ht!]
\caption{Performance evaluation of different backbones trained w/o DA on various datasets. \textit{Best} represents the highest performance achieved by different DA methods. \textit{Auto} indicts automatic  DA method, and \textit{Rand} means random DA.}
\label{table:advanced_DA}
\begin{adjustbox}{max width=0.95 \linewidth}
\begin{tabular}{llllll}
\toprule
DataSet      & Backbone  & No Aug & \textit{Best} &  Auto & Rand \\ \midrule
                                                   
                                                   
\multirow{3}{*}{PTB-XL}       & MLP       & 60.29 &66.34 & 62.77           & 61.86       \\ \cline{2-2}
                              & Conv-1D   & 75.00 &77.91 & 78.75           & 77.91       \\ \cline{2-2}
                              & ResNet-1D & 77.24 &78.03 & 78.21           & 79.60       \\ \midrule
                                                   
\multirow{3}{*}{Apnea-ECG}    & MLP       & 53.57 &57.90 & 57.42           & 57.84       \\ \cline{2-2}
                              & Conv-1D   & 80.14 &82.67 & 79.1            & 81.33       \\ \cline{2-2}
                              & ResNet-1D & 76.64 &77.74 & 73.52           & 77.64       \\ \midrule
                                                   
\multirow{3}{*}{Sleep-EDF}    & MLP       & 49.85 &56.27 & 53.68           & 55.38       \\ \cline{2-2}
                              & Conv-1D   & 83.62 &85.00 & 84.44           & 85.30       \\ \cline{2-2}
                              & ResNet-1D & 83.61 &85.01 & 85.26           & 85.16       \\ \midrule
                                                   
\multirow{3}{*}{MMIDB-EEG}    & MLP       & 77.24 &78.68 & 79.29           & 78.53        \\ \cline{2-2}
                              & Conv-1D   & 79.29 &79.72 & 81.23           & 81.66        \\ \cline{2-2}
                              & ResNet-1D & 76.16 &80.15 & 79.61           & 78.53        \\ \midrule
                              
\multirow{3}{*}{CLAS}         & MLP       & 76.71&79.06  & 74.93           & 74.91       \\ \cline{2-2}
                              & Conv-1D   & 62.83&77.96  & 73.26           & 68.73       \\ \cline{2-2}
                              & ResNet-1D & 76.54&76.54  & 74.65           & 69.36       \\ \midrule
                                                  
\multirow{3}{*}{PAMAP2}       & MLP       & 61.15&66.17  & 60.71           & 54.51       \\ \cline{2-2}
                              & Conv-1D   & 89.22&92.02  & 91.88           & 88.18       \\ \cline{2-2}
                              & ResNet-1D & 89.81&92.02  & 91.14           & 87.00       \\ \midrule
                                                  
\multirow{3}{*}{UCI-HAR}      & MLP       & 87.89&90.80  & 86.77           & 87.48       \\ \cline{2-2}
                              & Conv-1D   & 91.92&93.55  & 92.5            & 93.86       \\ \cline{2-2}
                              & ResNet-1D & 89.79&92.94  & 91.72           & 93.38      \\
    \bottomrule
\end{tabular}
\end{adjustbox}
\end{table}

\subsubsection{Performance evaluation of automatic and practical DA methods.}
The performance evaluation of automatic and random DA is displayed in Table.\ref{table:advanced_DA}, where both the baseline (\textit{No Aug}) and best performance achieved by different augmentations (\textit{Best}) from Table.\ref{table_different_DA} are also included for easy comparison. We can conclude that

I). \textit{Random DA is both effective and efficient} \textbf{(R3)}.
As we can see from Table.\ref{table_different_DA}, the performance trained with random DA generally outperforms the baselines (\textit{No Aug}), and are comparable with the best performance across different datasets and backbones.
Therefore, random DA is an efficient, effective, and practical automated data augmentation method for biobehavioral time series data.

II). \textit{Automatic DA outperforms most comparable baselines} (\textbf{R1}). From the various possible DA combinations and their corresponding magnitudes, the automatic DA method helps us search for the more effective DA policies (operations and magnitudes).
As we may find that the automatic DA method (\textit{Auto}) generally outperforms the baselines across different datasets and backbones, and achieves comparable or higher performance than the best performance, which is achieved by manually selected DAs. 

\subsubsection{Quantitative measurement of the effectiveness for different DA methods. (R4)} 
We tried to answer the question of why some DA methods are more effective than others, and how can we quantitatively measure the effectiveness of different DA methods?
First, we summarized the desired attributes for effective DAs:

1) \textit{challenging}: the augmented data should be complex and strong enough that a deep model must learn useful representations to perform the task. 

2) \textit{faithful}: the DA must not make the task impossible, being so strong that they destroy all features of the input. For example, the random noise added to jittering should not be so strong that makes learning impossible from the augmented data.

Next,   we adopted the metrics in\cite{gontijo2020tradeoffs} to quantify how DA improves model's performance by quantitatively measuring \textit{challenging} and \textit{faithful} for different DAs,.  As shown in Fig.\ref{fig:quantitative_evaluation} (c),  \textit{affinity} is used to measure how faithful of a DA, while \textit{diversity} measures the level of challenge resulting from applying an augmentation. 
Fig.\ref{fig:quantitative_evaluation} (a) and (b) measure both \textit{affinity} and \textit{diversity} across 65 different augmentations for PTB-XL and PAMAP2 dataset. We find that many augmentations that dramatically decrease the performance have low affinity and high diversity (upper left). On the other hand, many successful augmentations (\textit{blue points}) lay in the area with  high affinity, and for fixed (high) value of affinity, test accuracy generally increases  with the increase of diversity.  

In summary, \textit{affinity} and \textit{diversity} together provide an explanation of an augmentation policy's benefit to a model's performance. 
In other words, effective DA is a trade-off between generating more diverse data that should be more difficult for a model to fit, while remaining faithful that the learning task is still possible from the augmented data.

\begin{figure*}[ht!]
  \centering
  \subfloat[Number of operation]{\includegraphics[width=0.35\linewidth]{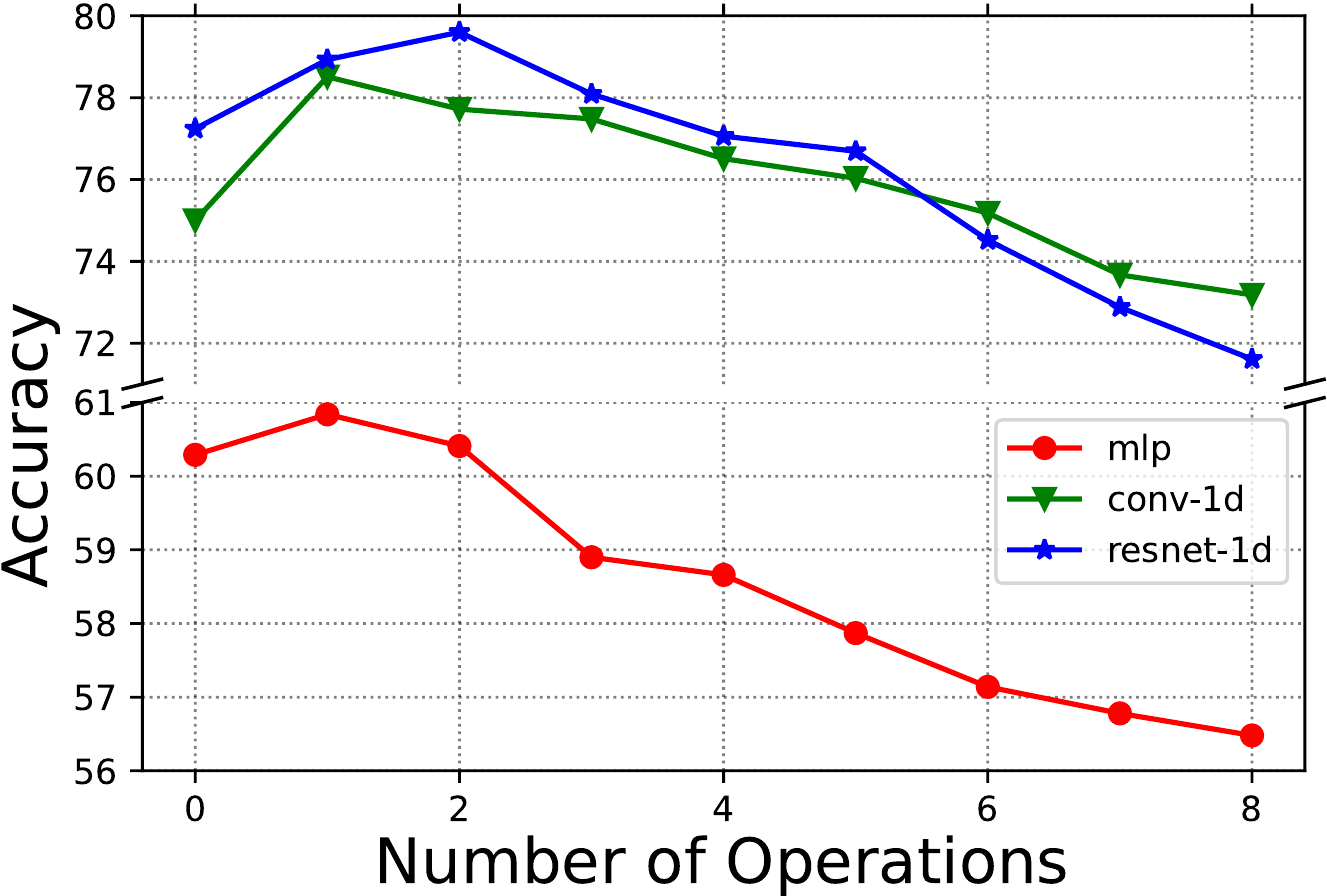}}
    \qquad
    \qquad
    \qquad
  \subfloat[Magnitude]{\includegraphics[width=0.35\linewidth]{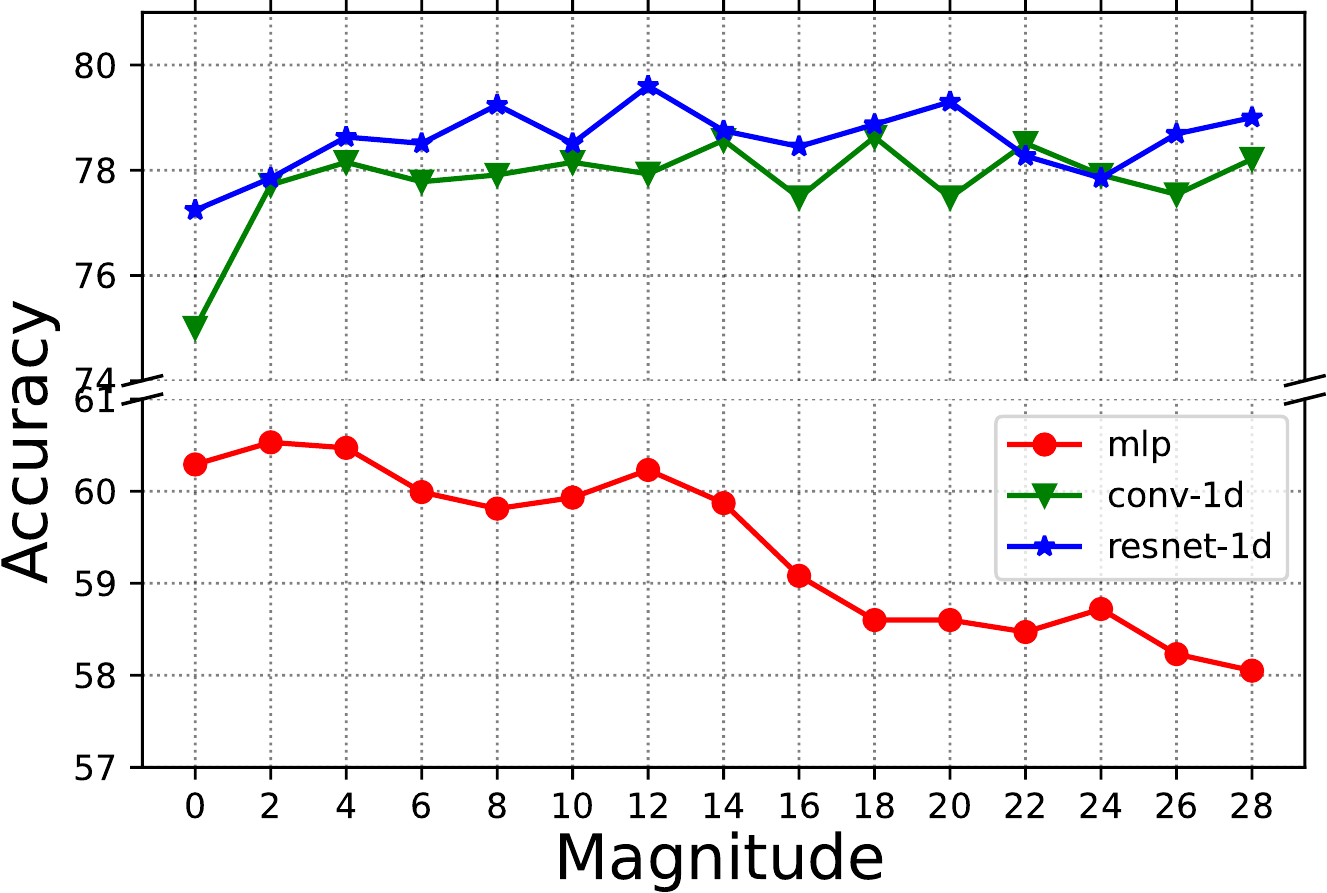}}
  \caption{Performance when number of operation (a) and magnitude (b) is changed on the PTB-XL dataset. }
  \label{fig:randaugment_MN}
\end{figure*}

\subsection{Ablation Study}

\subsubsection{The effect of different number of operations and different value of magnitude in random DA}
The random DA method achieves improved or comparable performances across different tasks and datasets using the fixed number of transformations and fixed value of magnitude.
To further study the sensitivity of random DA to the selection of transformations and magnitude,  we run experiments of random DA with different number of operations and magnitude on the PTB-XL dataset.  The results in Fig.\ref{fig:randaugment_MN} (a)  (\textit{fixed M}) suggest that the random DA improves performance as the number of operations is increased, even with only 1 operation ($J=1$). However, the performance is dropped once $J > 4$, which is potentially caused by the fact that the augmented data is too challenging for the model to learn anything meaningful after $J$ consecutive augmentations, therefore, leading to dropped performance.

A constant magnitude $M$ sets the distortion magnitude to a constant number during training. To validate the impact of different magnitudes,  we run experiments with fixed $J=2$ and varied numbers in terms of magnitude. The results in Fig.\ref{fig:randaugment_MN} (b) suggest that the model is not very sensitive to the change of magnitude (\textit{except MLP, as it is a relatively weak backbone}). That is because first, the magnitude is limited in a small reasonable range for all the operations (\textit{details in supplementary materials}); and second, the composition of $J$ random consecutive operations could potentially alleviate the sensitivity to the different magnitudes.



\begin{figure}[ht!]
  \centering
  \includegraphics[width=1.0\linewidth]{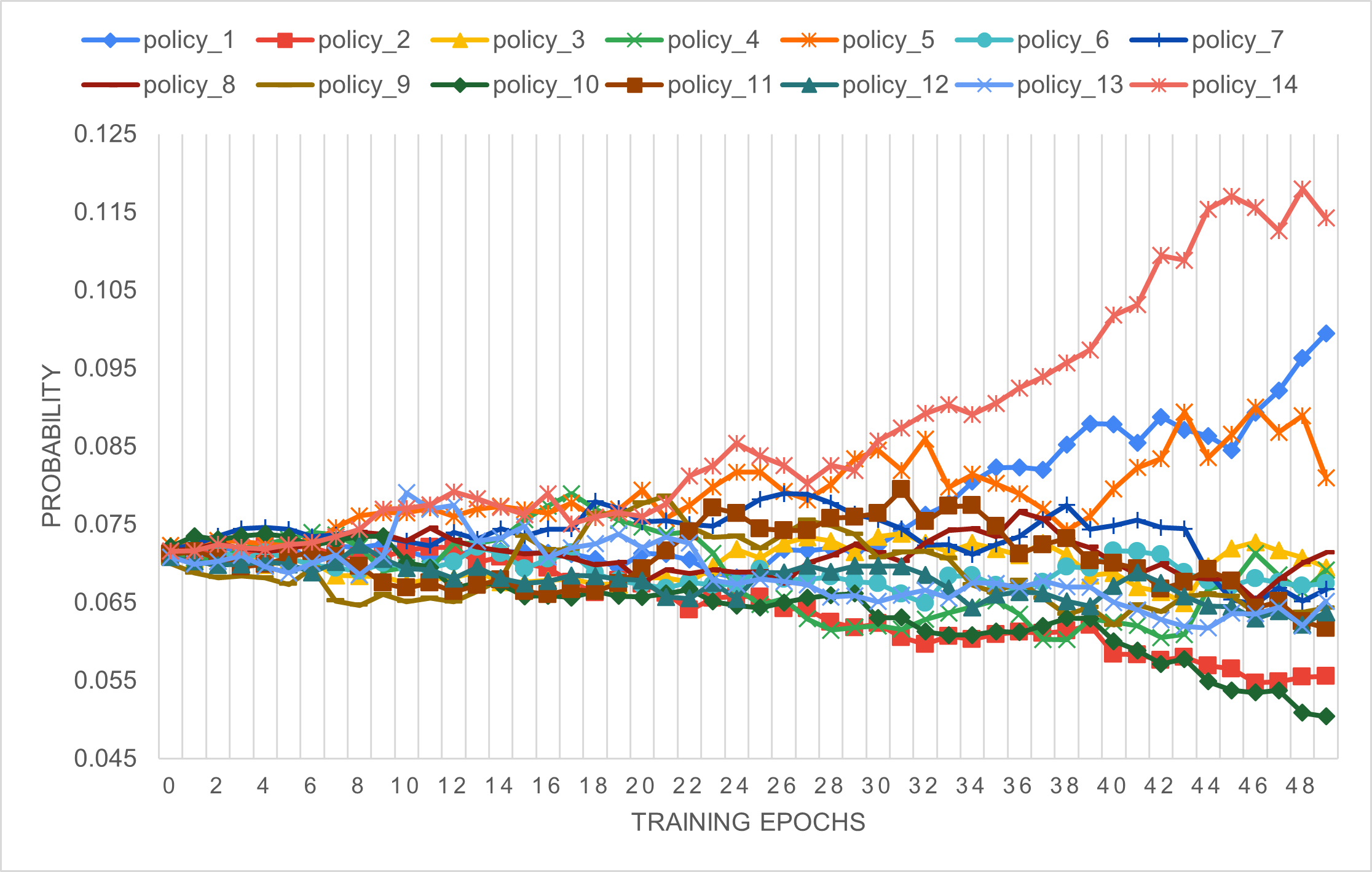}
  \caption{The development of probability parameters for each DA policy in the PTB-XL dataset. All policies start with the same probability, and are increased or decreased in terms of probability during training. Higher probability means the policy has much higher chance to be selected, indicting the corresponding policy is more effective.}
  \label{fig:DA_DADA_ops_weight_small}
\end{figure}

\subsubsection{Visualization of policies in automatic DA}
An illustration of the development of probability (\textit{normalized}) for individual DA policy over training epochs is shown in Fig.\ref{fig:DA_DADA_ops_weight_small}, where all the 14 sub-policies start from the same probability, and change with the training epochs.
For example, the probability is gradually increased for \textit{sub-policy\_1, sub-policy\_5} and \textit{sub-policy\_14}, indicting those policies are more effective than others.
Although automatic differentiable DA shows improved performance, it comes at a cost in terms of computational complexity, which is roughly four times than the corresponding baseline models.

\section{Conclusion}
In this work, we first conducted a comprehensive and systematic evaluation of various basic DAs on different biobehavioral  datasets, finding that all the datasets can benefit from DA if using them carefully. The effectiveness of an augmentation depended on both the dataset itself (\textit{data type, task}) and the used backbone.
Further,  we explored to adopt more recent DA techniques for biobehavioral signals by designing a different policy architecture. The results demonstrated that automatic DA can help learn  effective policies but at a cost of high computational complexity, while random DA was demonstrated to be both effective and efficient during our experiments.
At last, we attempted to answer the question of why DA is effective (\textit{or not}), by first summarizing two desired attributes (\textit{challenging} and \textit{faithful}), and then two we used two quantitative metrics  (\textit{diversity} and \textit{affinity}) to measure the corresponding attributes of DA. This can help understand why a DA works (\textit{or not}), and guide us for the search of more effective DA methods in the future.

We hope our experimental results could shed some light on DA for biobehavioral time series data, as a carefully selected DA could outperform many claimed improvements in the literature.  Our next plan is to investigate domain knowledge inspired DA, and also develop more efficient way to quantitatively measure the effectiveness of a DA method.

{\small
\bibliographystyle{ieee_fullname}
\bibliography{egbib}
}


\appendix

\Large{Appendix}
\normalsize
\section{Datasets}
\label{appendix:dataset}


\textbf{PTB-XL:} 
The PTB-XL \cite{wagner2020ptb} dataset is a large dataset containing 21,837 clinical 12-lead electrocardiogram (ECG) records from 18,885 patients of 10 second length, where 52\% are male and 48\% are female with ages range from 0 to 95 years (median 62 and interquantile range of 22).  
There are two sampling rates: 100 Hz and 500 Hz, available in the dataset, but in our experiments, only data sampled at 100 Hz are used.
The raw ECG data are annotated by two cardiologists into five major categories, including normal ECG (NORM), myocardial infarction (MI), ST/T Change (STTC), Conduction Disturbance (CD) and Hypertrophy (HYP).
The dataset contains a comprehensive collection of various co-occurring pathologies and a large proportion of healthy control samples. We experimented classifying all 5 cardiac conditions as learning tasks.
Further, to ensure a fair comparison of machine learning algorithms trained on the dataset, we follow the recommended splits of training and test sets, which results in a training/testing ratio of 8/1.

\textbf{Apnea-ECG:}
The Apnea-ECG \cite{penzel2000apnea} dataset studies the relationship between human sleep apnea symptoms
and heart activities (monitored by ECG). This database can be
accessed through Physionet\cite{goldberger2000physiobank}. This dataset contains 70 records with
a sampling rate of 100 Hz, from where 35 records were divided into training, and the other
35 were divided into the test set. The duration of the records varies from slightly less than
7 hours to nearly 10 hours. The labels were the annotation of each minute of each recording
indicating the presence or absence of sleep apnea. Thus, we split the ECG recording into
each minute, which was a total of 6000 data points for each separation. We extracted 17233
samples for the training set and 17010 samples for the test set. And the ratio of non-apnea
and apnea samples in the training set was 61.49\% to 38.51\%.

\begin{table*}[]
\centering
\caption{Details of datasets used in the experiments.}
\label{table:details}
\begin{adjustbox}{max width=\textwidth}
\begin{tabular}{llcclllcl}
\toprule
    Dataset & Data type & \# Subjects & \# Channels & \# Length & \# Train & \# Test & \# Classes & Task                             \\ \midrule
PTB-XL \cite{wagner2020ptb}  & ECG       & 18885       & 1           & 1000      & 14618    & 1652    & 5          & cardiac condition classification \\ \midrule
Apnea-ECG  \cite{penzel2000apnea}  & ECG       & 32          & 1           & 6000      & 17233    & 17010   & 2          & sleep apnea detection            \\ \midrule
Sleep-EDF  \cite{kemp2000analysis} & EEG       & 22          & 1           & 3000      & 31731    & 10577   & 5          & sleep stage recognition           \\ \midrule
MMIDB-EEG  \cite{Schalk2004}       & EEG       & 109        & 64           & 640      & 3708     & 927      & 2         & movement recognition  \\ \midrule 
CLAS  \cite{markova2019clas} & EDA       & 62          & 1           & 960       & 993      & 359     & 2          & stress and affect detection      \\ \midrule
PAMAP2  \cite{reiss2012introducing} & IMU       & 9           & 52          & 1000      & 4775     & 677     & 12         & human activity recognition       \\ \midrule
UCI-HAR \cite{reyes2016transition} & IMU       & 30          & 9           & 128       & 7352     & 2947    & 6          & human activity recognition       \\ 
\bottomrule
\end{tabular}
\end{adjustbox}
\end{table*}

\textbf{Sleep-EDFE:} 
The Sleep-EDF (expanded) \cite{kemp2018sleep} dataset contains whole-night sleep recordings from 822 subjects with physiological signals and sleep stages that were annotated manually by well-trained technicians. In this dataset, the physiological signals, including Fpz-Cz/Pz-Oz electroencephalogram (EEG), electrooculogram (EOG), and chin electromyogram (EMG), were sampled at 100 Hz. We targeted to detect 5 sleep stages. including wakefulness, stage N1, N2, N3, and REM \cite{berry2012rules}. To model the relationship between the sleep patterns and physiological data, we split the whole-night recordings into 30-second Fpz-Cz ECG segments as in \cite{supratak2020tinysleepnet}, which resulted in a total of 42308 ECG and sleep pattern pairs. We divided 25\% of the samples into a testing set according to the order of the subject IDs.

\textbf{MMIDB-EEG:} The MMIDB-EEG dataset \cite{Schalk2004} studies the relationship between physiological EEG and human body physical/imaginary movement. This dataset contains over 1,500 EEG recordings in 1-2 minutes with a sampling rate of 160Hz from 109 subjects. Each subject performed baseline (eyes open and close) and four tasks, including open and close left or right fist, imagine opening and closing left or right fist, open and close both fists or feet, and imagine opening and closing both fists or both feet. Following \cite{roots2020fusion}, we omit the data from 6 subjects due to the incorrect annotations and split the remaining data into 4s segments, which results in 4635 segments in total. Our task focuses on the classification between baseline and physical hand movement. Further, data from 22 subjects, based on the order of subject ids, is split into the test set, and the rest samples are employed as the training set.


\textbf{CLAS:} 
The CLAS dataset \cite{markova2019clas} aims to support research on the automated assessment of certain states of mind and emotional conditions using physiological data. The dataset consists of synchronized recordings of ECG, photopletysmogram (PPG), electrodermal activity (EDA), and acceleration (ACC) signals. There are 62 healthy subjects who participated and were involved in three interactive tasks and two perceptive tasks. The perceptive tasks, which leveraged the images and audio-video stimuli, were purposely selected to evoke emotions in the four quadrants of arousal-valence space. In this study, our goal was to use the EDA signal to detect binary high/low stress states that are annotated in arousal-valence space. We processed the raw EDA data with a lowpass Butterworth filter with a cutoff frequency of 0.2 Hz, then split the sequences into 10-second segments. We divided the train/test set in a subject-independent manner and utilized the data from 17 subjects as the test set according to subject ids ($>$ 45).

\textbf{PAMAP2:}
The PAMAP2 \cite{reiss2012introducing} physical activity monitoring dataset consists data of 18 different physical activities, including household activities (sitting, walking, standing, vacuum cleaning, ironing, etc)  and a variety of exercise activities (Nordic walking, playing soccer, rope jumping, etc), 
performed by 9 participants wearing three inertial measurement units (IMU) and a heart rate monitor.
Accelerometer, gyroscope, magnetometer and temperature data are recorded from the 3 IMUs placed on three different locations (1 IMU over the wrist on the dominat arm, 1 IMU on the chest and 1 IMU on the dominant side's ankle) with sampling frequency 100Hz.
Heart rate data are recorded from the heart rate monitor with sampling frequency 9Hz. 
The resulting dataset has 52 dimensions (3 x 17 (IMU) $+$ 1 (heart rate) = 52). 
Following the same setting as used in \cite{moya2018convolutional, tamkin2020viewmaker},
we linearly interpolated the missing data (upsampling the sampling frequency from 9Hz to 100Hz for heart rate), then took random 10s windows from subject recordings with an overlap of 7s, using the same train/validation/test splits.
As mentioned in  \cite{moya2018convolutional, tamkin2020viewmaker},  12 of the total 18 different physical activities are used in the experiments.

\textbf{UCI-HAR:}
The UCI-HAR \cite{reyes2016transition} dataset was collected from a group of 30 volunteers with an age range from 19 to 48 years. During the data collection, all the subjects wore a smartphone (Samsung Galaxy S II) with embedded inertial sensors around their waist and were instructed to follow an activity protocol  performing six basic activities, including three static postures (standing, sitting, lying) and three dynamic activities (walking, walking downstairs and walking upstairs). The dataset also included postural transitions that occurred between the static postures, but they were discarded in our experiments (also in related works), due to a much smaller size for those transitions.  3-axial linear acceleration and 3-axial angular velocity were captured using the embedded accelerometer and gyroscope of the smartphone at a constant rate of 50Hz. Following the authors' suggestion, the sensor signals were sampled in a fixed-width sliding windows of 2.56 second and 50\% overlap, resulting in 128 readings per window.

\begin{table*}[]
\centering
\caption{List of basic DA methods discussed in this paper. Additionally, the range of magnitude for individual operations is also reported. Some operations do not use the magnitude information (e.g. Rotation).}
\label{table_description}

\begin{adjustbox}{max width=1.0\textwidth}

\begin{tabular}{l l l  c}
\toprule
Operation Name & Description & \begin{tabular}[c]{@{}l@{}} Range of \\ Magnitude\end{tabular}   &  Default Value\\ \midrule
Jittering  & Add noise to the inputs  &  [0, 0.2]   & 0.03  \\  \midrule
Scaling    & \begin{tabular}[c]{@{}l@{}}changes the magnitude of the data in a window by \\ multiplying a random scalar.\end{tabular} &  [0, 0.5]   & 0.1 \\  \midrule
Rotation   & \begin{tabular}[c]{@{}l@{}}Rotate each element by a random rotation matrix.\\ A widely used alternative is flipping.\end{tabular}   &  N/A  & N/A \\  \midrule
Permutation & \begin{tabular}[c]{@{}l@{}}Perturb the location of the data in a single window\end{tabular} &   [0, 8]  &  5 \\ \midrule
Magnitude Warping & \begin{tabular}[c]{@{}l@{}}Changes the magnitude of each sample by \\ convolving the data window with a smooth curve.\end{tabular}& [0, 0.5]  &  0.2\\   \midrule
Time Warping  & \begin{tabular}[c]{@{}l@{}}Perturb the temporal location by smoothly \\ distorting the time intervals between samples.\end{tabular} & [0, 0.5]  & 0.2   \\       \midrule
Window Slicing &  \begin{tabular}[c]{@{}l@{}}Slice time steps oﬀ the ends of the pattern, which is \\ equivalent to cropping for image data augmentation.\end{tabular} &   [0.5, 1.]  & 0.9    \\\midrule
Window Warping  &\begin{tabular}[c]{@{}l@{}} takes a random window of the time series and\\ stretches it by 2 or contracts it by 1/2  \end{tabular} &  [0, 0.3]   & 0.1\\ 
\bottomrule
\end{tabular}
\end{adjustbox}
\end{table*}

\section{Quantitative Measurement for Data Augmentation}
In \cite{gontijo2020tradeoffs}, Affinity is defined as the ratio between the validation accuracy of a model trained on clean data and tested on  an augmented validation set, and the accuracy of the same model tested on clean data. More formally, let $\mathcal{D}_{train}$ and $\mathcal{D}_{val}$ be the training and validation datasets, and let $\mathcal{D}_{val}^{'}$ be an augmented dataset derived from $\mathcal{D}_{val}$. Further, let $\mathcal{F} ( \cdot )$ be a model trained on $\mathcal{D}_{train}$, and $\mathcal{A}_{cc}(\mathcal{F}, \mathcal{D})$ denote the accuracy of the model when evaluated on dataset $\mathcal{D}$. Then, for an augmentation $\tau$, the Affinity is defined as:
\begin{align} 
    Affinity (\tau) = \frac{\mathcal{A}_{cc}(\mathcal{F}, \mathcal{D}_{val}^{'})}{\mathcal{A}_{cc}(\mathcal{F}, \mathcal{D}_{val})}, \\
      \mathcal{D}_{val}^{'} = \big\{(\tau(x), y) | \forall(x, y) \in \mathcal{D}_{val} \big\}
\end{align}

Diversity is defined as the ratio of final training loss of a model trained with a given augmentation, and the loss of the model trained on original data.
Formally, let $\tau$ be an augmentation and $\mathcal{D}_{train}$ and $\mathcal{D}_{train}^{'}$  be the training data and augmented training data respectively.
Further, let $\mathcal{L}(\theta | \mathcal{D})$  be the training loss of a model with parameter $\theta$ on the training data $\mathcal{D}$. Then, we can define the Diversity as:
\begin{align}
    Diversity (\tau) = \frac{ \mathcal{L}(\theta | \mathcal{D}_{train}^{'})}{ \mathcal{L}(\theta | \mathcal{D}_{train})}, \\
    \mathcal{D}_{train}^{'} = \big\{(\tau(x), y) | \forall(x, y) \in \mathcal{D}_{train} \big\}
\end{align}

\section{Model Architecture}
A list of basic DA methods used in the paper is included in Table.\ref{table_description}, where the range of magnitude, and default value for individual DA is illustrated.
The detailed information of the Conv-1D and ResNet-1D are illustrated in Table. \ref{table:conv-1d} and Table. \ref{table:ResNet-1d}.

The detailed information of the MLP, Conv-1D and ResNet-1D are illustrated in Table.\ref{table:mlp_structure}, Table. \ref{table:conv-1d} and Table. \ref{table:ResNet-1d}.
\begin{table}[ht!]
\caption{Structure of the MLP model. \textit{B}: batch size; \textit{L}: length of sequence; \textit{C}: number of channels. }
\label{table:mlp_structure}
\begin{adjustbox}{max width=0.9\linewidth}
\small
\begin{tabular}{lllc}
\toprule
Layer Name & Input Shape   & Output Shape     & Parameter                              \\ \midrule
Reshape    & {[}B, L, C{]} & {[}B, L*C{]}    & -                                      \\ \midrule
Linear-1   & {[}B, L*C{]} & {[}B, 500{]}    & {[}L*C, 500{]}                         \\ \midrule
\multicolumn{4}{c}{BatchNorm, ReLU, Dropout}                  \\ \midrule
Linear-2   & {[}B, 500{]} & {[}B, 256{]}     & {[}500, 256{]}    \\ \midrule
\multicolumn{4}{c}{BatchNorm, ReLU, Dropout}           \\ \midrule
Linear-3   & {[}B, 256{]} & {[}B, Classes{]} & {[}256, Classes{]}\\ \bottomrule
\end{tabular}
\end{adjustbox}
\end{table}

\begin{table}[ht!]
\centering
\caption{Structure of the Conv-1D model. \textit{B}: batch size; \textit{L}: length of sequence; \textit{C}: number of channels.}
\label{table:conv-1d}
\begin{adjustbox}{max width=0.9\linewidth}
\small
\begin{tabular}{lllc}
\toprule
Layer Name & Input Shape        & Output Shape                & Parameter          \\ \midrule
Reshape    & {[}B, L, C{]}      & {[}B, C, L{]}               & -                  \\ \midrule
Conv1d-1  & {[}B, C, L{]}      & {[}B, 32, L{]}             & {[}1x5, 32{]}      \\ \midrule
\multicolumn{4}{c}{BatchNorm, ReLU, Max Pooling}           \\ \midrule
Conv1d-2   & {[}B, 32, L/3{]}   & {[}B, 64, L/3{]}            & {[}1x5, 64{]}      \\ \midrule
\multicolumn{4}{c}{BatchNorm, ReLU, Max Pooling}        \\ \midrule
Conv1d-3   & {[}B, 64, L/9{]}   & {[}B, 128, L/9{]}           & {[}1x5, 128{]}     \\ \midrule
\multicolumn{4}{c}{BatchNorm, ReLU, Max Pooling}    \\ \midrule
Conv1d-4  & {[}B, 128, L/27{]} & {[}B, 256, L/27{]} & {[}1x5, 256{]}     \\ \midrule
\multicolumn{4}{c}{BatchNorm, ReLU}     \\ \midrule
Average Pool    & {[}B, 256, L/27{]} & {[}B, 256, 1{]}            & -                  \\ \midrule
fc-1       & {[}B, 256{]}    & {[}B, Classes{]}            & {[}256xClasses{]} \\ \bottomrule
\end{tabular}
\end{adjustbox}
\end{table}

\begin{table}[ht!]
\centering
\caption{Structure of the ResNet-1D model. \textit{B}: batch size; \textit{L}: length of sequence; \textit{C}: number of channels.}
\label{table:ResNet-1d}
\begin{adjustbox}{max width=0.9\linewidth}
\small
\begin{tabular}{cccc}
\toprule
Layer Name                        & Input Shape                             & Output Shape                            & Parameter          \\ \midrule

Reshape                           & {[}B, L, C{]}                           & {[}B, C, L{]}                           & -                  \\  
Conv1d                            & {[}B, C, L{]}                           & {[}B, 64, L{]}                          & 1x3, 64,  max pool \\  \midrule
Layer1\_x                         & {[}B, 64, L/2{]}                        & {[}B, 64, L/2{]}                        &    $ \begin{bmatrix}
1\times3, & 64\\
1\times3, & 64 
\end{bmatrix} \times 2$       \\  \midrule 
Layer2\_x                         & {[}B, 64, L/2{]}                        & {[}B, 128, L/4{]}                       &     $ \begin{bmatrix}
1\times3, & 128\\
1\times3,& 128 
\end{bmatrix} \times 2$                \\  \midrule
Layer3\_x  & {[}B, 128, L/4{]} & {[}B, 256, L/8{]}  &   $ \begin{bmatrix}
1\times3, & 256\\
1\times3,& 256 
\end{bmatrix} \times 2$     \\ \midrule
Layer4\_x   & {[}B, 256, L/8{]}  & {[}B, 512, L/16{]} &    $ \begin{bmatrix}
1\times3, & 512\\
1\times3,& 512 
\end{bmatrix} \times 2$      \\ \midrule
Average pool & {[}B, 512, L/16{]} & {[}B, 512, 1{]}    & -                  \\ \midrule
FC         & {[}B, 512{]}      & {[}B, Classes{]}   & {[}512, Classes{]} \\ \bottomrule
\end{tabular}
\end{adjustbox}
\end{table}

\end{document}